\documentclass{article} 
\usepackage[final]{colm2026_conference}

\usepackage{microtype}
\usepackage{hyperref}
\usepackage{url}
\usepackage{booktabs}

\usepackage{graphicx}
\usepackage[most]{tcolorbox}

\usepackage{lineno}

\definecolor{darkblue}{rgb}{0, 0, 0.5}
\definecolor{sagegreen}{RGB}{138,154,91}
\hypersetup{colorlinks=true, citecolor=darkblue, linkcolor=darkblue, urlcolor=darkblue}

\title{When Contextual Inference Fails: Cancelability in Interactive Instruction Following}

\author{
 \textbf{Natalia Bila\textsuperscript{2,*}},
 \textbf{Kata Naszádi\textsuperscript{1,*}},
 \textbf{Alexandra Mayn\textsuperscript{2}},
 \textbf{Christof Monz\textsuperscript{1}},
\protect\\
  \textsuperscript{1}Language Technology Lab, University of Amsterdam, \protect\\
  \textsuperscript{2}Department of Language Science and Technology, Saarland University
\protect\\
  {\small * These authors contributed equally to this work. \textbf{Correspondence:} \href{mailto:k.naszadi@uva.nl}{k.naszadi@uva.nl}}
}

\begin{document}

\ifcolmsubmission
\linenumbers
\fi

\maketitle

\begin{abstract}

We investigate the separation of literal interpretation from contextual inference in a collaborative block-building tasks, where an agent must resolve underspecified instructions using context. We adapt an existing two-speaker psycholinguistic paradigm into an interactive benchmark called Build What I Mean (BWIM). This setup contrasts a pragmatically cooperative speaker with one who is only literally reliable. In BWIM, models face underspecified instructions and must choose between making a contextual inference or requesting clarification at a small communication cost. Evaluating several state-of-the-art LLMs, we find a clear dissociation between judgment and action. Although models successfully detect speaker unreliability in explicit confidence ratings, they fail to leverage this awareness when taking action. Instead of deploying efficient clarification strategies, models default to suboptimal behaviors. These include partner-blind over-clarification and question-averse guessing under uncertainty. BWIM provides a controlled environment to evaluate online partner adaptation and contextual reasoning in interactive settings.
\end{abstract}

\section{Introduction}
Many linguistic theories posit that human language processing relies on a two-tier system consisting of a more general literal meaning and pragmatics -- the latter being responsible for a flexible contextual meaning creation building on top of literal meaning \citep{levinson2000presumptive,sperber1986relevance,grice1975logic}. To what extent is such a two-tier system beneficial for artificial conversational agents? And are literal and pragmatic processing already represented in conversational systems? This paper investigates these questions through a recently introduced task designed to test whether an agent can distinguish between literal meaning and contextual inference.

\begin{figure}

    \centering
    \includegraphics[scale=0.8]{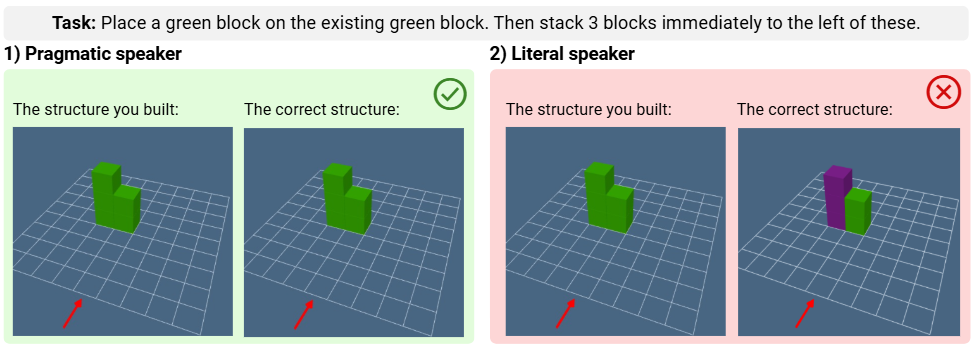}
    \caption{Example trial with an underspecified instruction ("stack 3 blocks") that requires the builder to infer the color of the blocks. The default pragmatic inference is to continue with the same color as in the previous step. In the Pragmatic speaker condition (left), such default inferences are always correct. In the Literal speaker condition (right), this inference might lead to a mismatch with the target structure.}
    \label{fig:L0_feedback}
\end{figure}

A standard diagnostic for distinguishing literal meaning from pragmatic inference is cancellability: contextual inferences are defeasible and may fail in contexts where the literal content remains reliable \citep{grice1975logic, levinson1983pragmatics}. In repeated interaction, this defeasibility can be revealed implicitly, as interlocutors learn that drawing contextual inferences with a given partner often leads to incorrect interpretations, whereas reliance on literal meaning does not. The task used in this paper leverages this asymmetry by testing whether conversational agents adapt their interpretations over time, selectively discounting contextual inferences while preserving literal content.
The task involves a building scenario where an agent interacts with two distinct speakers.
\begin{itemize}
    \item Speaker A expects the agent to draw standard contextual inferences.
    \item Speaker B is ``literal-only"; their instructions can only be trusted at the level of literal content, and drawing inferences might lead to errors (Figure ~\ref{fig:L0_feedback}).
\end{itemize}

Through repeated interaction, the agent must learn to trust or cancel inferences based on the speaker's identity. We refer to this process as \textit{in-context adaptation}: the agent accumulates interaction history across trials and changes its behavior according to the feedback. This differs from \textit{in-context learning} \citep{brown2020language}, which relies on examples provided in the prompt. Here, the agent draws conclusions about the speaker based on the mismatch between their default response and the provided feedback. This setup allows us to observe whether the agent maintains a representation of what was literally said versus what was inferred based on the context.

A further goal of this work is methodological. Much prior work on pragmatic competence in language models relies on explicit metalinguistic judgments, e.g., asking models to select or explain the intended interpretation of an utterance \citep{hu-etal-2023-fine, park-etal-2024-multiprageval, sravanthi2024pub, sravanthi-etal-2025-understand}. While informative, such evaluations probe pragmatic knowledge in isolation rather than its use in situated actions. In contrast, our task evaluates the ability to contextualize meaning implicitly, through task success and information-seeking behavior in an interactive setting. This allows us to assess whether models can deploy pragmatic inferences appropriately when they matter for achieving goals, rather than merely articulating them when prompted.

We evaluate three state-of-the-art LLMs on this task, comparing their performance against human benchmarks. Our evaluation proceeds in two stages. In the first, agents report confidence ratings after each trial, allowing us to track adaptation as shifts in subjective certainty and to assess whether models distinguish literal from contextually enriched interpretations. In the second, we introduce \textbf{Build What I Mean (BWIM)}, a new interactive benchmark that reframes the task as a multi-turn dialogue in which agents may seek clarification at a cost, allowing us to examine whether models translate uncertainty into effective information-seeking behavior.

Across our evaluations, we observe a clear dissociation between explicit and implicit ways of handling of uncertainty. Models often detect speaker unreliability in their confidence ratings, yet they fail to use this signal to produce efficient information-seeking strategies. Ultimately, our experimental design tests more than whether models can draw pragmatic inferences. It also reveals whether they can selectively cancel them based on partner-specific evidence. Finally, it shows whether this latent sensitivity appears in action, not just in judgment.

\section{The adaptation task}

\begin{table}
  \centering
  \renewcommand{\arraystretch}{1.6}
  \begin{tabular}{p{2.3cm} p{9.7cm}}
    \hline
    \textbf{Condition} & \textbf{Example Instruction} \\
    \hline
    Underspecified (number) & Stack three green blocks behind the existing green block. Build \textbf{a yellow stack} to the right of the green one. \\
    Underspecified (color)  & Stack three blue blocks in front of the existing blue blocks. Then stack \textbf{two blocks} to the left of the tower you just built. \\
    Fully-specified         & Stack three red blocks in the bottom right corner. Put \textbf{two yellow blocks} on top of the red stack you just built. \\
    \hline
  \end{tabular}
  \caption{\textbf{Example items for the two experimental conditions: fully-specified and underspecified instructions.} \textit{Fully-specified} instructions include information about the color and number of building blocks. In the \textit{underspecified condition}, either color, or number of blocks is omitted, requiring models to infer the intended information.}
  \label{tab:example_items}
\end{table}

The setup used to test whether agents are aware of contextual inferences involves two speaker profiles in a block-building game. The task is to construct simple structures on a 3D grid that satisfy instructions given by two types of agents: Literal Lisa or Pragmatic Pia. Instructions from both speakers are sometimes underspecified, omitting the color or the number of blocks to be used (Table ~\ref{tab:example_items}). Pragmatic Pia shortens her utterances only when the missing information can be recovered from context, whereas Literal Lisa occasionally makes errors by omitting information even when the contextual inference leads to an incorrect interpretation (see Figure \ref{fig:L0_feedback}).

The relevant contextual inference is that, if not stated otherwise, the color and size of a structure should match those of the most recently built one. This inference has been observed in previous dialog-based block-building games \citep{narayan-chen-etal-2019-collaborative,pmlr-v220-kiseleva23a,mohanty2022collecting}. For example, in an interaction from the Minecraft dialog corpus \citep{narayan-chen-etal-2019-collaborative}, an architect instructs the builder to add another block to a completed row without specifying its color. The builder resolves this ambiguity by inferring that the color matches the most recently used block, but the architect cancels this inference in the next turn by providing more information. Computational models trained on these datasets have been shown to reproduce this inference \citep{naszadi-etal-2023-aligning}. It was also shown in a human study with the same stimuli that participants draw these inferences spontaneously without prior exposure to the task \citep{naszadi2026listeners}.  This ability to automatically draw contextual inferences allows the builder agent to learn about the partner’s reliability from feedback obtained during repeated interactions with the same speaker. After each structure is submitted, the builder receives feedback indicating the target structure that the speaker intended to communicate.  As illustrated in Figure \ref{fig:L0_feedback}, a human participant inferred the intended color, assuming the speaker was pragmatic, but the contextual inference was revealed to be wrong in the feedback.

Throughout this paper, we use \emph{pragmatic competence} in a broad sense, referring to the contextualization of meaning beyond what is explicitly stated, rather than to specific pragmatic phenomena such as scalar implicatures or Gricean maxims in isolation. In our task, pragmatic reasoning consists in resolving underspecified instructions using shared contextual assumptions and selectively canceling such enrichments when they prove unreliable.

After each structure, the builder is asked to report, on a 4-point scale, how confident they are that they built the structure intended by the speaker. A builder that can distinguish literal content from contextual inferences is expected to maintain high confidence on fully specified trials, while gradually lowering confidence when following underspecified instructions from Literal Lisa.

Human data from the confidence-scoring version of this task already supports the existence of rapid adaptation. However, we are ultimately interested in whether models can translate their uncertainty into clarification questions to guarantee task success. To test this, we converted the task into the \textbf{Build What I Mean} (\textbf{BWIM}) benchmark. While the instructions remain identical to the confidence-based adaptation task, models in \textbf{BWIM} have the option to ask questions and receive answers. This allows them to learn the utility of information-seeking in an interactive environment. Because asking questions incurs a small cost, we expect models to behave efficiently by asking fewer questions when interacting with Pragmatic Pia than with Literal Lisa.

The human behavioral data used for comparison in this paper are taken from \citep{naszadi2026listeners}. Here, we use a subset of these results solely as a benchmark for model behavior, without revisiting the authors' cognitive modeling analyses.

\section{Related work}

  Pragmatic inference may be implemented in multiple ways in computational models. The first group of models builds in the separation of literal and pragmatic meaning by design. In these approaches, there is a literal model responsible for the first-pass interpretation and a separate module responsible for further contextualization of the meaning. One of the most prominent models in this line is the Rational Speech Acts framework \citep{goodman2016pragmatic}, which models pragmatic inference as Bayesian inference about the partner’s production process on top of initial literal lexical definitions.

This model has been computationally operationalized in many tasks, including reference games \citep{white2020learning}, navigation tasks \citep{fried2018speaker}, and human-robot collaboration for goal assistance \citep{tan2024pragmatic}. It has further been extended to research in language learning \citep{Monroe2015LearningIT, mcdowell2019learning, naszadi2023communicating}.
These modular architectures allow for the targeted handling of pragmatic inferences, as the component responsible for enrichment is architecturally distinct from the literal foundation.

The second group of models attempts to learn pragmatic behaviors directly from data. These often utilize end-to-end neural architectures where pragmatic nuances are captured within the latent representations of large language models (LLMs). While some works confirm emergent pragmatic abilities in LLMs \citep{hu-etal-2023-fine, park-etal-2024-multiprageval}, others have highlighted significant gaps in this area, suggesting that large language models may not be pragmatic enough in zero-shot settings \citep{liu2023we, lipkin2023evaluating, kurch2024large, ruis2023goldilocks, srikanth-etal-2024-pregnant}. To address these limitations, recent methodologies have been proposed to improve LLM pragmatic performance, such as incorporating communicative objectives during training \citep{sato-etal-2025-pragmatic} or leveraging specialized prompting techniques \citep{sravanthi-etal-2025-understand, Fardin+25:Gricean}. In this paper, we investigate what happens when pragmatic inferences drawn by such end-to-end models go wrong and whether agents can selectively cancel them.

Most existing evaluations of pragmatic competence in LLMs rely on explicit interpretation tasks, such as multiple-choice questions, entailment judgments, or direct prompts asking models to explain or select an intended meaning \citep[e.g.,][]{hu-etal-2023-fine, park-etal-2024-multiprageval, sravanthi2024pub, halat-atlamaz-2024-implicatr, kimyeeun-etal-2024-developing}. Some evaluations use generative settings, assessing models based on their ability to produce contextually appropriate responses reflecting non-literal processing \citep{yerukola-etal-2024-pope, srikanth-etal-2024-pregnant}, though these still primarily evaluate one-shot responses rather than the integration of pragmatic inferences into goal-directed behavior. Our work focuses on implicit, task-oriented evaluation, where pragmatic success is measured through adaptation over time and action choices.

In interactive and visually grounded settings, agents often need to ask clarification questions to resolve ambiguous instructions. Similar to our approach, prior work uses a partner model to compute effective clarification strategies in reference games: however, this partner model remains static \citep{khalid-etal-2020-combining}. Other studies evaluate clarification requests in grounded tasks using uncertainty estimates based on predictive log-probabilities \citep{naszadi-etal-2023-aligning, testoni-fernandez-2024-asking}. These log-probabilities align poorly with human clarification behavior unless constrained to limited multiple-choice setups \citep{naszadi-etal-2023-aligning, allen-knowNo-2023}. More recent work explores calibration and uncertainty in multimodal reference resolution, showing that current models remain poorly calibrated when deciding whether clarification is needed \citep{ali-etal-2026-reference}. In our work, the source of uncertainty is strictly controlled. Disambiguation is context-driven across extended grounded interactions, requiring a partner model that adapts over time.
\section{Experiments}

We report two experiments investigating how LLMs adapt to different speakers' reliability. In Experiment 1, after each trial, models rated their certainty that they built a correct structure, which allowed to detect adaption via changes in confidence ratings. In Experiment 2, models were given points for building correctly and penalized for errors. On each trial, they could ask clarification questions at a small cost. Thus, adaptation was measured via changes in question-asking behavior across speaker blocks.

Each speaker block contained 20 instructions: 8 of them were fully specified (unambiguous) and 12 were underspecified, omitting information about the number or color of blocks (6 items of each type of underspecitfication). After completing the task, the builder receives feedback about whether the built structure matched the target structure. In the Pragmatic Pia condition, feedback was always consistent with pragmatic interpretation. In the Literal Lisa condition, feedback was sometimes pragmatic ($n = 4$) and more often literal ($n = 8$), allowing builders to infer that contextually licensed pragmatic interpretations were not always correct. Retaining a subset of pragmatically consistent trials prevented the literal speaker from appearing deliberately uncooperative, forcing agents to adapt to her reliability rather than abandoning pragmatic reasoning altogether or inventing alternative communication rules.

\textbf{Models} We evaluated three LLMs: Gemini 2.5 Pro \citep{comanici2025gemini}, GPT 5.1 \citep{openai_gpt5_1_system_card_2025}, and Claude Opus 4.5 \citep{anthropic_claude_opus4_5_system_card_2025}. The models were selected based on their performance during the pilot phase: their ability to correctly interpret fully specified instructions and respond in the specified format, and sensitivity to ambiguity (lower ratings in the critical trials compared to unambiguous control trials). 
We excluded DeepSeek-R1 \citep{guo2025deepseek} based on low accuracy on unambiguous trials and failure to respond in the specified format. Two additional models (Llama-3-70B-Instruct \citep{llamaHerd2024}  and Grok-4.1 \citep{xai2026grok4}) were evaluated during the pilot but were not included in the main experiment because while their building accuracy was comparable to other models, their confidence ratings were consistently at ceiling, showing no sensitivity to the experimental manipulation. 

\textbf{Task adaptation for LLMs} In the original experiment by \cite{naszadi2026listeners}, human participants read textual instructions and responded by building structures with colored blocks in a 3D space. As feedback, they received a message indicating whether the response was correct and a  picture of the intended structure.

In our experiment, this task was converted into a fully textual format. Models were instructed to respond with coordinates and colors of the blocks they would use. We created a main task prompt that described the grid including (1) the coordinates of the origin and each corner; (2) valid x, y, and z values; and (3) the required response format ("Color:x;y;z"). As feedback, they received a message indicating whether the response was correct, the coordinates they sent, and the target coordinates (see Appendix~\ref{diagram} for the full session structure). The full prompt and examples of a response and feedback are provided in Appendix \ref{main_prompt}.

The final set of models (openai/gpt-5.1, google/gemini-2.5-pro, anthropic/claude-opus-4.5) was queried via hosted APIs at default temperature. 

\subsection{Confidence rating experiment}

At the beginning of each session, models received a system prompt describing the study context (similar to what human participants saw). Each prompt contained the description of the grid, building instructions, and a request to rate their certainty (on a scale of 1 to 4). After each round, the model's response was appended to the chat history and followed by a feedback message revealing the intended target structure. Appending the history is crucial as it provides learning signals for adaptation. Each main block contained instructions from one speaker (Literal Lisa or Pragmatic Pia), with a transition message indicating a speaker change. At the end of the experiment, the model was asked to comment on the differences in instruction style between the two speakers.

Each session contained 40 instructions in total (20 instructions per speaker block) and was run as an independent conversation, with no shared context across sessions. 

To ensure that results reflect sensitivity to instruction type and speaker feedback, rather than artifacts of specific instructions, item order, or speaker sequencing, we created 8 experimental lists with counterbalanced item sets and speaker order. The order of the instructions was pseudorandomized such that underspecified instructions appeared in the same position across lists. As a result, each model was assessed across 30 sessions, each using a distinct sequence of items. Reported results are averaged across all sessions per model.

\subsection{Interactive QA experiment: Build What I Mean}

In this experiment, instead of giving certainty ratings, models could resolve ambiguity by asking clarification questions. The responses were generated by gpt-4o-mini and were consistently correct brief answers (e.g., "3 blocks" or "Yellow"). Models were awarded $+10$ points for building a correct structure and penalized for building incorrectly ($-10$ points). They could reduce the probability of losing points by asking one question per trial, which incurred a $-5$ points cost\footnote{This interactive version of the benchmark is available at \href{https://github.com/ltl-uva/build_what_i_mean}{github.com/ltl-uva/build\_what\_i\_mean}.}. This cost was added to obtain an informative measure of model behavior: in the pilot experiment, when clarification questions carried no cost, models asked questions on almost every underspecified trial, making it impossible to distinguish between genuine uncertainty and a default strategy of always asking for clarification. An example interaction between a model and the pragmatic speaker is presented in Figure ~\ref{fig:exp2-example}. The procedure and materials were otherwise identical to Experiment 1. There were 4 lists with a counterbalanced speaker order.

\begin{figure}[t]
\centering
\begin{tcolorbox}[colback=sagegreen!10,colframe=sagegreen!80,title=Example Interaction]

\textbf{Pia:} Initial structure: Blue,0,50,0;Blue,-100,50,0;Blue,100,50,0. Behind the rightmost blue block, build a red stack of three blocks. Build a yellow stack directly to the right of the red one.

\textbf{Gemini:} How high should the yellow stack be?\\  
\textbf{Pia:} 3 blocks high (-5 points for asking).

\textbf{Gemini:} Blue,0,50,0; Blue,-100,50,0;\\ Blue,100,50,0; Red,100,50,-100;\\ Red,100,150,-100; Red,100,250,-100; \\Yellow,200,50,-100; Yellow,200,150,-100; Yellow,200,250,-100.

\textbf{Pia:} Correct structure built! (+10 points) Round score: +5. Total score: +25.

\end{tcolorbox}
\caption{An example \textbf{BWIM} interaction. The session starts with a system prompt with a description of the grid and the required answer format, then in each round, the speaker sends building instructions. The model has an option either to respond directly or to ask a clarification question at a cost. After receiving the model's final answer, the speaker sends feedback and updates the score. Thus, the model must balance the risk of building incorrectly and the cost of asking questions in order to maximize the final score.}
\label{fig:exp2-example}
\end{figure}

\section{Results}
\subsection{Unambiguous trials} On fully-specified trials, all models showed mean accuracy $>  0.95$ and mean ratings $> 3.85$ (out of 4), with no difference between speaker blocks. This means that models are able to perform the building task and have high confidence and good calibration when encountering fully detailed instructions.

\subsection{Pragmatic Inference}
All models demonstrated the expected pragmatic inference without prior examples or explicit instructions on resolving under-specification. This confirms zero-shot pragmatic competence across all tested models. When interacting with Pragmatic Pia, the models resolved underspecification pragmatically in the vast majority of cases: GPT (98\%), Claude (92.4\%), and Gemini (88.8\%).

In the Literal Lisa condition -- where models frequently encounter pragmatic errors -- the agents continued to lean toward contextually plausible interpretations but began incorporating `noisy' actions. These actions suggest a strategy of guessing or randomizing responses when faced with a pragmatically unreliable partner. This mirrors human behavioral data: while some people maintain a pragmatic bias and merely lower their confidence, others begin to randomize their interpretations in an attempt to optimize outcomes against an inconsistent speaker \citep{naszadi2026listeners}. Specifically, when feedback suggested a pragmatic error was likely, Claude maintained a pragmatic choice in 92.4\% of cases, while GPT dropped to 79\% and Gemini to 70\%. An analysis of the responses that differed from the contextually licensed interpretation can be seen in  the Appendix~\ref{non-prag}.
\begin{figure*}[t]
  \includegraphics[width=\linewidth]{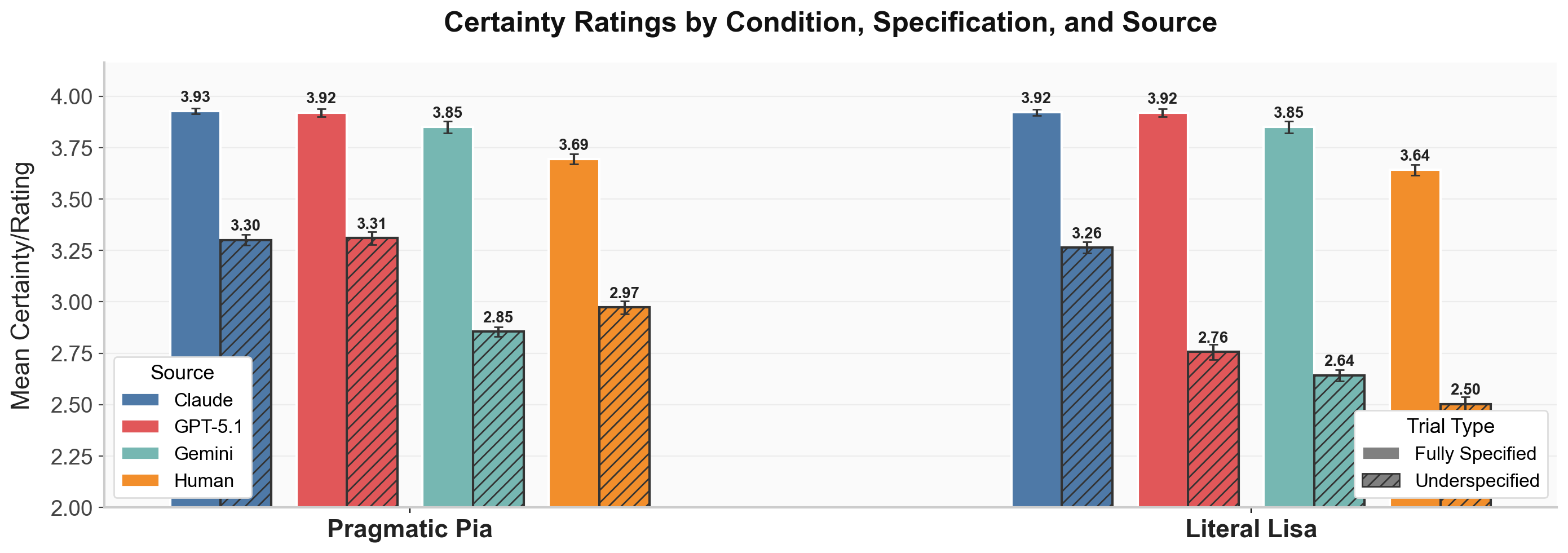}
  \caption{\textbf{Mean certainty ratings by condition, specification, and source.} Bars show mean certainty ratings for pragmatic and literal speaker across three models and human participants, distinguishing between fully-specified and underspecified trials. Error bars indicate standard error. All models and human participants gave higher ratings to fully-specified trials in both speaker conditions. Underspecified trials were rated higher in the pragmatic condition.}
  \label{fig:res_combined}
\end{figure*}
\subsection{Effect of speaker identity} As illustrated in Figure \ref{fig:res_combined}, the effect of speaker identity is selectively localized to underspecified trials. Across all models, performance and confidence remained high for fully specified instructions, regardless of whether the partner was Pragmatic Pia or Literal Lisa. This indicates that models correctly attribute the source of error to underspecification rather than general communicative incompetence; they continue to fully trust literal content in both conditions.

The divergence between fully specified and underspecified trials serves as a metric for how models notice pragmatic violations. A larger gap in the Literal Lisa condition suggests that the model recognizes the partner's pragmatic unreliability. While this gap is most pronounced in human participants, it remains statistically significant for Gemini and GPT. In contrast, Claude appears relatively insensitive to partner identity, maintaining similar behavior across both conditions.

A regression analysis showed that speaker identity did not affect ratings on unambiguous trials. In contrast, there was a significant effect of speaker identity on underspecified trials: all models gave lower ratings in the literal condition, however, the magnitude of the adaptation effect differed across models. GPT and Gemini showed a stronger adaptation effect with a difference in ratings of 0.6 and 0.28 points between the two speaker conditions, respectively. Claude was the least sensitive to speaker identity with difference in ratings of only $0.1$ points. Full model output is included in Appendix~\ref{app:stats}.

\subsection{Adaptation over time}
In the results reported so far we ignored the fact that accumulation of information about the partner unfolds over time: the builder learns about the partner's reliability through repeated interaction. In Figure \ref{fig:res_combined} we show confidence ratings coupled with interpretation type (pragmatic vs.\ other) for 4 time segments for each partner. We see similar patterns as before: Claude is not sensitive to partner type, while Gemini and GPT both show patterns of adaptation. Gemini employs a guessing strategy heavily when talking to Literal Lisa and does not recover from this fully when switching to new partner. GPT shows the strongest adaptation effects with progressively lower confidence choices and more guessing with the erroneous literal partner and rapid adaptation and high confidence in the pragmatic inferences when the partner is also reliably pragmatic.

\begin{figure*}[t]
  \includegraphics[width=\linewidth]{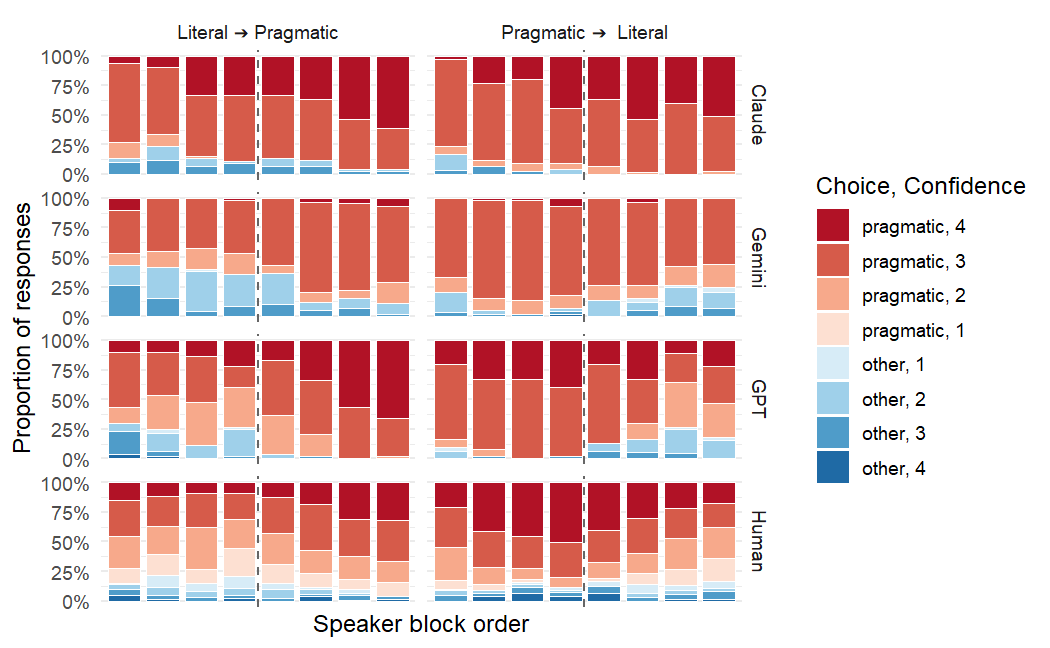}
  \caption{\textbf{Confidence and response change throughout interactions with different speakers.} Proportion of pragmatic and non-pragmatic responses are shown for each model and human participant and speaker order (Literal Lisa or Pragmatic Pia). Colors indicate response type and confidence rating (1-4 scale). Vertical dashed lines indicate the change of speaker. Each speaker block is divided into 4 time segments. GPT's adaptation effects mirror those of human participants, while Gemini shows a strong carry-over of mistrust that was atypical in human participants and Claude shows no sensitivity to partner type.}
  \label{fig:adaptation}
\end{figure*}

\subsection{Clarification questions} 
The paradigm of recording confidence effectively captures bottom-up adaptation from task-oriented feedback. However, replacing explicit ratings with clarification questions provides a more natural, implicit signal of uncertainty. In this \textbf{BWIM} setting, an adaptive agent should ideally modulate its information-seeking behavior based on the perceived reliability of its partner and minimize costs with a pragmatic speaker while increasing inquiries with a literal one.

As shown in Table \ref{tab:clarification_results}, however, the adaptive behavior of the tested models is significantly less robust in this interactive environment than their performance on confidence rating might suggest. These results reveal a big difference in how models handle the trade-off between communication cost and task success. GPT-5.1 exhibits question-aversion, asking almost no questions across the entire session. While this minimizes communication costs, it caps accuracy at 73.1\%.

\begin{table}[h]
\centering
\begin{tabular}{lrrr}
\hline
\textbf{Model} & \multicolumn{1}{c}{\textbf{\#Q (Pia)}} & \multicolumn{1}{c}{\textbf{\#Q (Lisa)}} & \multicolumn{1}{c}{\textbf{Acc.}} \\
\hline
GPT & 0.25 & 0.00 & 73.1\% \\
Gemini & 11.75 & 12.00 & 85.6\% \\
Claude Opus & 8.50 & 9.50 & 89.4\% \\
\hline
\end{tabular}
\caption{Average clarification questions per speaker block and overall task accuracy.}
\label{tab:clarification_results}
\end{table}

\begin{table}[h]
\centering
\begin{tabular}{lccc}
\hline
\textbf{Speaker} & \textbf{Total number} & \textbf{\% of trials} \\
\hline
Literal & 297 & 37.1\% \\
Pragmatic & 220 &  27.5\% \\
\hline
\end{tabular}
\caption{Clarification questions per speaker block asked by human participants (n = 40, 50\% of the total sample who asked at least one question). Unlike the models in Table~\ref{tab:clarification_results}, humans received no response and no benefit from asking, yet they still asked more questions in the Literal speaker condition, indicating the sensitivity to speakers' pragmatic reliability that the models failed to show, despite having the actual benefit from asking.}
\label{tab:clarification_results_humans}
\end{table}

In contrast, Gemini 2.5 Pro and Claude Opus 4.5 achieve significantly higher accuracy by adopting an inquisitive strategy. Both models correctly localize their uncertainty, with nearly all questions occurring during underspecified trials. However, both models remain partner-blind. Gemini asks an almost identical number of questions from both Pragmatic Pia  and Literal Lisa, and Claude shows only a marginal increase when facing Lisa. The models do not realize that they could save costs when playing with Pia by relying on her pragmatic reliability; instead, they treat every underspecified instruction as a trigger for clarification regardless of the source. This contrasts with human participants' behavior (see Table ~\ref{tab:clarification_results_humans} ) who showed sensitivity to speakers' communicative style and asked more questions in the literal speaker condition, even though, for humans, asking a questions did not incur an explicit penalty and required only the effort of typing.

\subsection{Models' feedback about the two speakers} 
All models noticed that Pia's instructions were easier to follow. However, they mentioned that Lisa omitted information more frequently than Pia, despite both speakers having an equal number of underspecified trials. This suggests that models detected the difference between the two speakers but explained it in terms of omitted information rather then reliability of inferences. For some example answers see  Appendix \ref{app:feedback}.

\section{Discussion}

These results provide an initial probe into whether language models distinguish between what is literally said and what is inferred from context. Two of the three models were sensitive to the speaker condition in their confidence scores, but only when the utterance triggered a contextual inference and not on the unambiguous control trials. This suggests that their representations are sensitive to the distinction between literal content and contextual inference. This is particularly interesting given that models are trained on pragmatically enriched language, where meaning routinely goes beyond literal content. The present paradigm therefore offers a first step toward asking how strongly models represent the distinction between literal and inferential meaning. Such a distinction may also be important for error correction: misunderstandings of literal content and misunderstandings of contextual inference require different kinds of repair, with the latter potentially requiring models to revise their assumptions about the speaker or context rather than the linguistic content itself.

The contrast between explicit confidence ratings and clarification behavior highlights the diagnostic value of implicit, task-oriented evaluation. Although some models exhibited sensitivity to pragmatic unreliability in their confidence judgments, this uncertainty did not reliably translate into effective information-seeking strategies in the \textbf{BWIM} setting. Instead, we observed a distinct dissociation. Models either engaged in high-confidence guessing and avoided even low-cost clarification, or incurred unnecessary communication costs by querying a reliable partner rather than leveraging pragmatic inference.

Pragmatic understanding in deployed agents requires acting under uncertainty, deciding when to infer and when to ask, and adapting to partner-specific behavior over time. The proposed task offers a concrete step toward evaluating these abilities in a controlled manner and provides a foundation for studying contextualized meaning creation in language-based situated agents.

\section{Limitations}

The contextual enrichment tested in this work is intentionally highly controlled. Each trial consists of one or two sentences establishing context followed by a single target instruction, with varied but still very constrained lexical realization. This design allows us to isolate pragmatic inference and its cancellability with high precision, but it limits the linguistic richness of the context. In natural multi-turn instruction-following, contextual cues for inference may arise from a wider range of discourse phenomena (e.g., anaphora, incremental updates, or expressions such as “another” or “one more”), which are not present in our materials.

In addition, the \textbf{BWIM} benchmark uses a fixed, explicit cost for clarification questions. This does not mirror the variable or implicit costs of communication in real-world interaction.

\section*{Acknowledgments}
We thank Vera Demberg and John Duff for their input on this project. Sergey Troshin and Evgeniia Tokarchuk helped with the implementation of the \textbf{BWIM} benchmark using the \textit{Agent2Agent} protocol.

This work was supported by the European Research Council (ERC) under the European Union’s Horizon 2020 Research and Innovation Programme (ERC Starting Grant ``Individualized Interaction in Discourse'', Grant Agreement No.~948878).  This research was also funded in part by the Netherlands Organization for Scientific Research (NWO) under project number VI.C.192.080. We also received funding from the Hybrid Intelligence Center, a 10-year programme funded by the
Dutch Ministry of Education, Culture and Science through the Netherlands Organisation for Scientific Research with grant number 024.004.022.

\bibliography{custom}
\bibliographystyle{colm2026_conference}

\appendix

\section{Experimental session structure}

\begin{figure}

    \centering
    \includegraphics[scale=0.4]{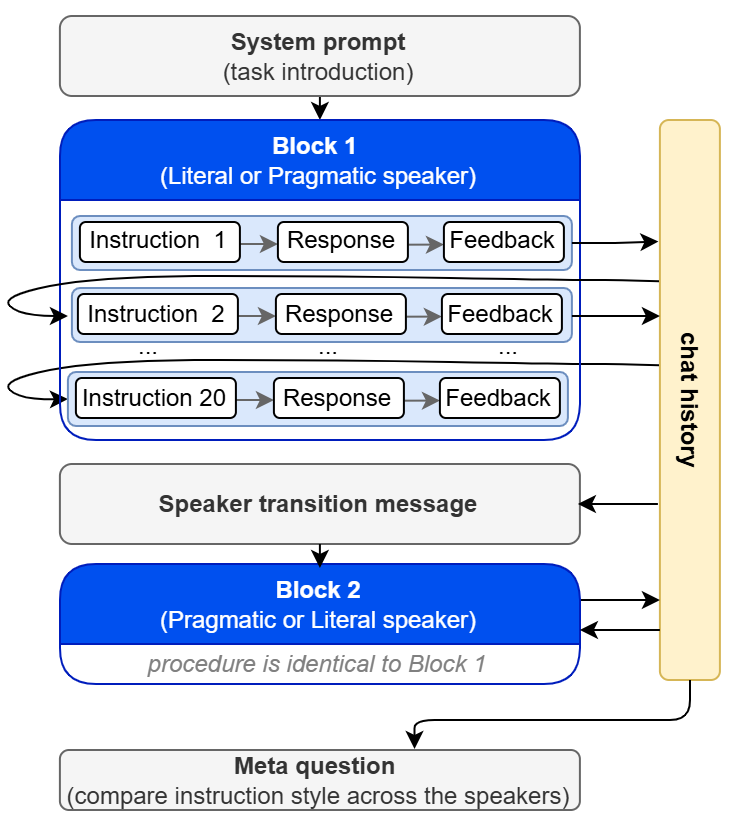}
    \caption{Structure of a single experimental session.}
    \label{fig:diagram}
\end{figure}

Both experiments share the same session structure (Figure~\ref{fig:diagram}). Each block starts with a system prompt describing the task. In Block 1, models follow the instructions from one speaker (Literal or Pragmatic). Each trial (the instruction, model's response, and feedback) is appended to the conversation history, such that each subsequent trial contains information about the previous ones. After the transition message, the same procedure is repeated with another speaker. At the end of the session. models are asked to compare the instruction styles of the two speakers.

The two experiments differ only at the trial level. In Experiment 1, models respond with the coordinates of the structure and a confidence rating. In Experiment 2, models either respond with the coordinates, or ask a clarification question before sending the coordinates.

\label{diagram}

\section{Prompts}
\label{main_prompt}
\subsection{Confidence rating experiment}

System prompt:
\begin{quote}
 \ttfamily{We are interested in studying how people write and follow instructions. 

We recently conducted another online study where participants wrote instructions of how to build structures in a simple 3D block building interface. 

In this study, your task will be to follow the instructions sent by two randomly selected previous participants and build the described structures. 

First, you will follow instructions written by one participant. Then you will follow instructions written by another participant. 

Now, you are going to do two practice trials and then follow the instructions written by \{speaker\}.}
\end{quote}

Main task prompt:
\begin{quote} \ttfamily{
CONTEXT:\\ 
Grid: 9x9 cells. Origin="middle square": center (0,0), is highlighted. The grid is the x–z plane. In front of you is the bottom left corner (-400,0,400) and the bottom right corner (400,0,400). Top right corner is (400,0,-400), top left corner is (-400,0,-400).\\ 

Valid x,z:[-400,-300,-200,-100,0,100,200,300,400]. Y(ground)=50; each extra block adds +100; valid y values are [50,150,250,350,450]. The grid may or may not contain an existing structure. "Existing structure: nan" means that the grid is empty. "A stack" means more than one block. \\

Output:"Coordinates:Color,x,y,z; Color,x,y,z;Rating:"; items separated by ";"; no spaces; write coordinates of all blocks that are on the grid, including the initial coordinates; color should be capitalized.\\
 
YOUR TASK: \\
Existing structure: \{start\_structure\}. \{task\}.\\
On a scale of 1-4, rate how certain you are that this is the structure that the previous participant saw, 1 means 'not certain at all', 4 means 'very certain'.\\

YOUR ANSWER:}
\end{quote}

Response example: 

\begin{quote}
\ttfamily{Coordinates: Purple,100,50,0; Purple,100,150,0; Purple,100,250,0; Purple,200,50,0; Purple,300,50,0;\\
Rating:3}
\end{quote}

Feedback example:
\begin{quote}
    \ttfamily{FEEDBACK: False;\\ 
    the structure you built = Purple,100,50,0; Purple,100,150,0; Purple,100,250,0; Purple,200,50,0; Purple,300,50,0;\\
    the correct structure = Purple,100,50,0; Purple,100,150,0; Purple,100,250,0; Green,200,50,0; Green,300,50,0;}
\end{quote}

\subsection{Interactive QA experiment}

Main task prompt:
\begin{quote}
\ttfamily{
Grid: 9x9 cells. Origin="middle square": center (0,0), is highlighted. The grid is the x–z plane. In front of you is the bottom left corner (-400,0,400) and the bottom right corner (400,0,400). Top right corner is (400,0,-400), top left corner is (-400,0,-400).\\

Valid x,z: [-400,-300,-200,-100,0,100,200,300,400]. Y(ground)=50; each extra block adds +100; valid y values are [50,150,250,350,450]. The grid may or may not contain an existing structure. The grid might be empty.\\ 

Output: "Coordinates: Color,x,y,z; Color,x,y,z;" items separated by ";"; no spaces; write coordinates of all blocks that are on the grid, including the initial coordinates; color should be capitalized. Only one question is allowed.\\ 

Scoring: Building the correct structure earns +10 points. Asking a question costs -5 points. Building an incorrect structure costs -10 points.}
\end{quote}

\section{Analysis of nonpragmatic responses.} 
\label{non-prag}
\begin{figure}[t]
  \centering
  \includegraphics[scale=0.5]{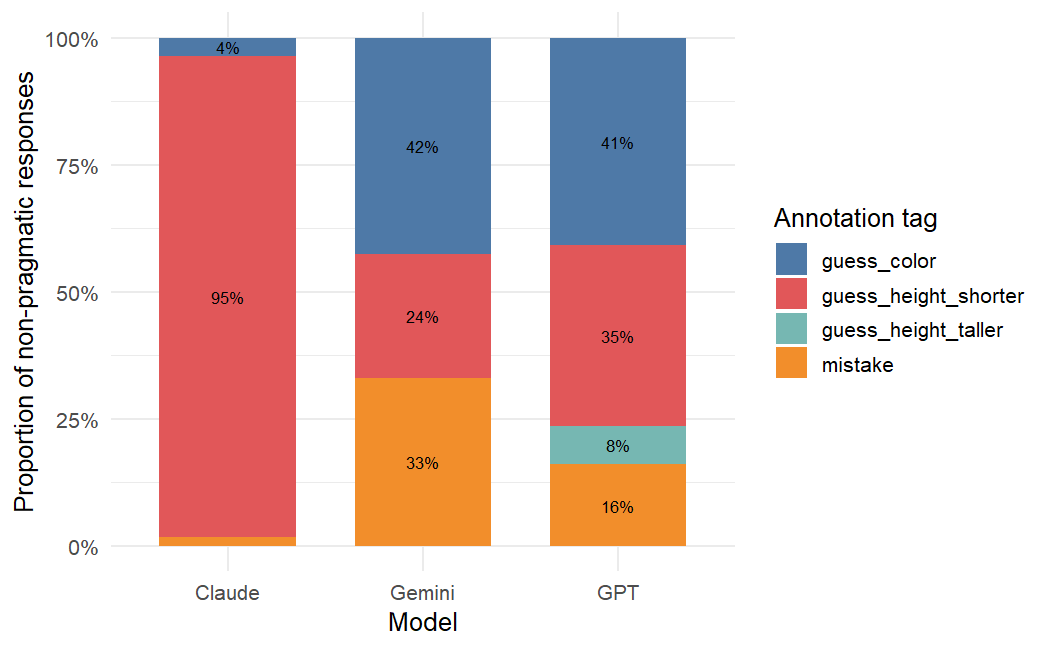}
  \caption{\textbf{Distribution of non-pragmatic response types across models.} Responses were annotated as \textit{mistake}  (e.g.: misplaced blocks) and as \textit{guesses} when models try to make a random guess to resolve the omitted information. The \textit{guess} category was divided by type: different color, shorter or taller height of stacks. The result suggests that models differed in ambiguity-resolution strategies, with Claude guessing mainly height, while Gemini and GPT tried to guess both height and color.}
  \label{fig:nonprag-annotation}
\end{figure}

Responses that differed from the contextually licensed interpretation were annotated as \textit{mistakes} when models misinterpreted instructions and  as \textit{guess} when models tried to come up with a resolution for the underspecification by randomly picking the color or number blocks. The nonpragmatic response rate differed across models, see Figure \ref{fig:nonprag-annotation}. 

\section{Models' responses about the speakers}
\label{app:feedback}

Table \ref{tab:feedback} shows answers from the models' feedback about the speakers. This represents well that the models did not attribute the source of difference to pragmatics when explicitly asked about it.
\begin{table}[htb]
  \centering
  \begin{tabular}{p{1cm} p{11cm}}
    \hline
    \textbf{Model} & \textbf{Feedback} \\
    \hline
    Claude & \textbf{Pia was more consistent about specifying colors} throughout their instructions.
\textbf{Lisa sometimes assumed I would infer colors or stack heights from context}, which occasionally led to misunderstandings
Both participants struggled occasionally with specifying exact quantities when using the word "stack" without a number. \\
    Gemini  & Pia was a much better instructor. \textbf{Their instructions were more complete}, which led to fewer errors and a higher degree of confidence on my part. Following Lisa felt like a guessing game at times, while following Pia felt more like a puzzle of careful interpretation. \\
    GPT      & Both participants were understandable in terms of spatial relations, but \textbf{Lisa’s  instructions were more ambiguous about colors and heights}, while Pia’s instructions were more explicit and internally consistent, making them easier to follow without guessing. \\
    \hline
  \end{tabular}
  \caption{Example feedback from models comparing instruction style of two speakers (Pragmatic Pia and Literal Lisa). All models evaluated the pragmatic speaker's instructions as more consistent and easier to follow, while the literal speaker's instructions required more inference. Despite both speakers having an equal number of underspecified instructions, the literal speaker was perceived as omitting information more frequently.}
  \label{tab:feedback}
\end{table}

\section{Statistical analysis}
\label{app:stats}
We fit a linear mixed-effects regression model predicting the confidence ratings on each trial with model (Claude, GPT, Gemini; dummy-coded with Claude as reference), speaker (Literal or Pragmatic; dummy-coded with Literal as reference) and their interaction as fixed effects and a random intercept for item.

The intercept corresponds to the reference model (Claude) in the literal condition. Regression coefficients ($\beta$) represent changes in ratings relative to the baseline. The coefficient for the speaker condition corresponds to the change in ratings for Claude in the pragmatic condition. Coefficients for the interactions with speaker condition refer to the additional shift relative to the reference model in the pragmatic condition.

Confidence ratings were sensitive to speaker condition (Literal vs. Pragmatic) only on underspecified trials, with higher ratings in the Pragmatic speaker condition. On unambiguous trials, confidence ratings did not differ between speaker conditions.

Regression model summaries for underspecified and unambiguous trials are reported in Table ~\ref{tab:critical_coef} and Table ~\ref{tab:filler_coef}, respectively.

\begin{table}[h]
\centering
\caption{Regression coefficients for underspecified trials.}
\label{tab:critical_coef}
\begin{tabular}{lrrrr}
 \hline
Predictor & $\beta$ & SE & $t$ & $p$ \\
 \hline
Intercept & 3.245 & 0.053 & 60.98 & $< .001$ \\
Gemini & -0.622 & 0.040 & -15.40 & $< .001$ \\
GPT & -0.495 & 0.040 & -12.26 & $< .001$ \\
Speaker (Prag.) & 0.109 & 0.050 & 2.20 & .028 \\
Gemini $\times$ Prag. & 0.184 & 0.057 & 3.22 & .001 \\
GPT $\times$ Prag. & 0.501 & 0.057 & 8.75 & $< .001$ \\
 \hline
\end{tabular}
\flushleft
\footnotesize
$\beta$ = regression coefficient, SE = standard error
\end{table}

\begin{table}[h!]
\centering
\caption{Regression coefficients for unambiguous (fully-specified) trials.}
\label{tab:filler_coef}
\begin{tabular}{lrrrr}
 \hline
Predictor & $\beta$ & SE & $t$ & $p$ \\
 \hline
Intercept & 3.920 & 0.070 & 56.04 & $< .001$ \\
Gemini & -0.071 & 0.022 & -3.28 & .001 \\
GPT & -0.004 & 0.022 & -0.18 & .857 \\
Speaker (Prag.) & 0.007 & 0.022 & 0.32 & .750 \\
Gemini $\times$ Prag. & -0.004 & 0.031 & -0.14 & .887 \\
GPT $\times$ Prag. & -0.009 & 0.031 & -0.29 & .776 \\
 \hline
\end{tabular}
\flushleft
\footnotesize
$\beta$ = regression coefficient, SE = standard error
\end{table}

\section{Questions asked by models in the Interactive QA experiment}

Across all runs, nearly all clarification questions targeted the omitted information (color of number of blocks to use). Only four questions out of 168 total clarification questions (aggregated across all three models) were position-related and occurred on trials with comparatively difficult structures (e.g., "In which direction should I extend the longer base (towards the front or back)?"). Given the small number (2.4\%) of position-related questions, the examples below focus on questions that targeted omitted information on underspecified trials.\\

\textbf{Color omission}
\begin{quote}
    \textbf{Instruction}: Starting from the highlighted square, place a horizontal row of three yellow blocks going towards the left side of the grid. Stack two blocks on top of the leftmost block.\\
    
\textbf{Gemini}: What color are the two blocks stacked on the leftmost block?\\
\textbf{GPT}: What color should the two blocks stacked on top of the leftmost yellow block be?\\
\textbf{Claude}: What color should the two blocks stacked on top of the leftmost block be?\\
\end{quote}

\textbf{Number omission}
\begin{quote}
\textbf{Instruction}: Stack four green blocks to the right of the highlighted middle square. Stack blue blocks to the right of the green ones.\\

\textbf{Gemini}: How many blue blocks should be stacked?\\
\textbf{GPT}: How many blue blocks should be in the stack to the right of the green ones (should it also be 4 blocks tall like the green stack)?\\
\textbf{Claude}: How many blue blocks should be in the blue stack?
\end{quote}

\end{document}